\documentclass[11pt]{article}

\usepackage[final]{acl}

\usepackage{times}

\usepackage{latexsym}

\usepackage[T1]{fontenc}

\usepackage[utf8]{inputenc}

\usepackage{microtype}
\usepackage{multirow} 

\usepackage{inconsolata}

\usepackage{graphicx}
\usepackage{CJKutf8}

\usepackage{tikz}
\usetikzlibrary{arrows.meta, positioning, calc, shadows, backgrounds, fit}
\usepackage{CJKutf8}
\usepackage{amsmath}

\title{\textit{Jiuge-Tuiqiao}: An Interpretable Human-AI System for Classical Chinese Poetry Refinement}

\author{
  \textbf{Yufeng Han}\textsuperscript{1,2,3\thanks{Equal contribution.}},
  \textbf{Lifan Deng}\textsuperscript{1,2,3,4\footnotemark[1]},
  \textbf{Cunliang Kong}\textsuperscript{1,2,3}, \\
  \textbf{Wenhao Li}\textsuperscript{1,2,3},
  \textbf{Xin Cong}\textsuperscript{5},
  \textbf{Yuzhuo Bai}\textsuperscript{1,2,3},
  \textbf{Kangyang Luo}\textsuperscript{1,2,3},
  \textbf{Maosong Sun}\textsuperscript{1,2,3,6\thanks{Corresponding author.}}\\
  \textsuperscript{1}Department of Computer Science and Technology, Tsinghua University, Beijing  \\
  \textsuperscript{2}Beijing National Research Center For Information Science And Technology \\  
  \textsuperscript{3}Institute for Artificial Intelligence, Tsinghua University, Beijing \\
  \textsuperscript{4}Rixin College, Tsinghua University\
  \textsuperscript{5}Department of Statistics and Data Science, Tsinghua University\\
  \textsuperscript{6}Jiangsu Collaborative Innovation Center for Language Ability, Jiangsu Normal University, Xuzhou\\
  \texttt{hanyufeng@mail.tsinghua.edu.cn}, \texttt{dlf24@mails.tsinghua.edu.cn} \\
 }

\begin{document}
\maketitle
\begin{abstract}
Classical Chinese poetry composition has long valued \textit{Tuiqiao}, the iterative refinement of words, imagery, and prosody. However, many current AI poetry systems follow a one-shot generation paradigm, which reduces users to prompt providers and weakens their creative agency. We present \textit{Jiuge-Tuiqiao}\footnote{You can access our services at \url{https://github.com/jiangli-va/Jiuge-Tuiqiao}.}, an interactive human-AI collaborative system for classical Chinese poetry composition. The system is designed around a triadic model: user-driven control, ancient-guided evidence, and AI-assisted generation. Users can lock characters or lines, receive real-time prosody feedback, and obtain interpretable refinement suggestions grounded in high-frequency collocations, PPL-ranked classical lines, and structured knowledge extracted from classical encyclopedias. This design turns AI from an autonomous generator into a background assistant that supports the user's own process of poetic refinement. Preliminary experiments and user feedback indicate that \textit{Jiuge-Tuiqiao} provides controllable refinement mechanisms, traceable literary evidence, and a positively received interactive experience for classical Chinese poetry composition.
\end{abstract}

%
%

\section{Introduction}

Classical Chinese poetry composition emphasizes \textit{Tuiqiao}: a constrained iterative refinement where poets balance semantic precision, tonal prosody, and aesthetic coherence. The well‑known case of \textit{Jia Dao} deliberating between “push the moonlit gate”(\textit{Seng Tui Yue Xia Men}) and “knock the moonlit gate”(\textit{Seng Qiao Yue Xia Men}) shows how poets weigh linguistic accuracy, emotional nuance and formal integrity. Therefore, \textit{Tuiqiao} sharpens poetic language, and the process of \textit{Tuiqiao} embodies the creator’s subjectivity — including linguistic competence, cultural literacy, and stylistic voice. 

Although one-shot generation is useful for rapidly producing a draft, it offers limited support for users who wish to preserve selected local decisions, inspect literary evidence, and iteratively revise the remaining text:

\begin{itemize}
    \item \textbf{Lack of a \textit{Tuiqiao} process}: Existing ``one-click generation'' paradigms generate complete poems from prompts but offer no mechanism for users to make precise revisions to unsatisfactory local elements. In particular, many one-shot workflows return a complete poem in response to a prompt but do not allow users to preserve selected local decisions while regenerating the remaining positions.
    \item \textbf{Lack of traceable evidence}: Many model-based systems present candidate words without traceable literary or linguistic evidence, making it difficult for users to understand why a suggestion fits the context or to verify it against classical sources.
    \item \textbf{Lack of creator autonomy}: Most critically, many model-centered workflows do not give users complete decision-making authority over how a poem evolves. Users may provide prompts or select outputs, but cannot determine at every refinement step which characters or lines should be preserved, revised, regenerated, accepted, or rejected. This may weaken users' sense of authorship and make the interaction closer to content consumption than to creative composition.
\end{itemize}

\begin{figure}[t]
\centering
\resizebox{\columnwidth}{!}{
\begin{tikzpicture}[
    node distance=3.5cm,
    main node/.style={circle, draw=black, thick, fill=gray!5, text width=1.8cm, align=center, font=\small\bfseries, minimum size=2.2cm},
    rel arrow/.style={-{Stealth[scale=1.2]}, thick, bend left=25, shorten >=2pt, shorten <=2pt},
    label font/.style={font=\scriptsize\itshape, color=black!70}
]

    \node[main node] (user) {User\\(Driven)};
    \node[main node, below left=of user] (ancient) {Ancients\\(Guided)};
    \node[main node, below right=of user] (ai) {AI\\(Assisted)};

    \path[rel arrow] (user) edge node[left, xshift=-0.1cm, label font] {Keyword/Query} (ancient);
    \path[rel arrow] (ancient) edge node[right, xshift=0.1cm, label font] {Literary Clues} (user);

    \path[rel arrow] (ancient) edge node[below, yshift=-0.2cm, label font] {Knowledge Enhancement} (ai);
    \path[rel arrow] (ai) edge node[above, yshift=0.2cm, label font] {Retrieved Evidence} (ancient);

    \path[rel arrow] (ai) edge node[right, xshift=0.1cm, label font] {Candidates} (user);
    \path[rel arrow] (user) edge node[left, xshift=-0.1cm, label font] {Locks/Refinement} (ai);

\end{tikzpicture}
}
\caption{The collaborative workflow of \textit{Jiuge-Tuiqiao}. It shows the relationship where the User maintains creative subjectivity, Ancient Knowledge provides knowledge guidance, and AI offers generative assistance.}
\label{fig:triadic_model}
\end{figure}
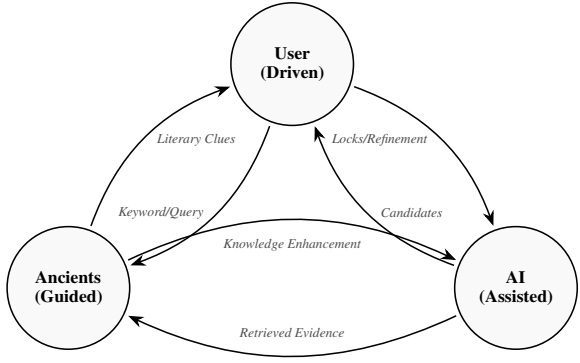

To bridge the aforementioned gaps, we present \textit{Jiuge‑Tuiqiao}. The system returns creative initiative to the poet by modeling a deep tripartite interaction among the user, the ancients, and AI (Fig\ref{fig:triadic_model}):
\begin{itemize}
    \item \textbf{User‑driven}: Users retain full control, can lock any satisfying character or line, and receive revision suggestions under those constraints — enabling true iterative refinement.
    \item \textbf{Ancient-guided}: A rich knowledge base (high-frequency collocations, famous lines and cues from ancient encyclopedias) provides literary or linguistically grounded suggestions, realizing ``refinement following the ancients' thinking''.
    \item \textbf{AI‑assisted}: AI steps to the background as a powerful language generation engine and knowledge retrieval tool, optimizing its outputs based on user feedback and knowledge constraints, rather than replacing the user’s creative agency.
\end{itemize}

We also integrate a dynamic prosody checking mechanism that provides real‑time feedback on tonal patterns and rhyme using color coding, helping users attend to both formal prosody and semantic expression during the \textit{Tuiqiao} process.

\textit{Jiuge-Tuiqiao} is fully implemented and accessible via a WeChat mini-program, open to poetry enthusiasts, teaching, and cultural dissemination, offering a complete workflow from generation, refinement to feedback. Preliminary user ratings indicate positive perceptions of agency, prosody guidance, trustworthiness, and overall satisfaction. Through this digital workflow, the system carries forward the traditional wisdom of \textit{Tuiqiao} in the AI era.

%
%
\section{Related Work}

\textbf{Classical Chinese Poetry Generation.} 
Automated poetry generation has evolved from rule-based methods to neural architectures optimizing coherence and style \citep{yi-etal-2018-chinese, yang-etal-2018-stylistic, yi-etal-2018-automatic}. Recent research leverages Large Language Models (LLMs) for structural and imagery control, including unified GPT-2 frameworks \citep{hu-sun-2020-generating}, imagery-focused masking in PoemBERT \citep{huang-shen-2025-poembert}, token-free character control in CharPoet \citep{yu-etal-2024-charpoet}, and domain-specific LoRA fine-tuning \citep{xie-2025-system}.

\textbf{Interactive Poetry Systems.} 
Moving beyond one-shot generation, interactive systems increasingly support human--AI collaboration in poetry composition. \textit{Jiuge} \citep{guo-etal-2019-jiuge} enables repeated revision and dynamic regeneration of unsatisfactory parts, while \textit{Yu Sheng} \citep{ma-etal-2023-yu} provides constrained generation and fine-grained polishing. Building on these interactive approaches, \textit{Jiuge-Tuiqiao} further makes the evidence behind refinement suggestions visible to users. It combines character- and line-level locking and real-time prosodic feedback with traceable evidence from high-frequency collocations, attributed historical lines, and ancient encyclopedia entries, allowing users to inspect retrieved evidence while retaining final control over each revision decision.

\textbf{Interactive AI in Other Domains.} 
Interactive refinement and human-in-the-loop feedback have also enhanced argumentative writing training \citep{ding-etal-2025-feat} and narrative coherence in storytelling \citep{martins-etal-2025-upon}. We extend this philosophy of collaborative, feedback-driven intervention to the highly constrained domain of classical Chinese poetry.

%
%
\section{System Architecture}

We formalize \textit{Jiuge-Tuiqiao} as a human-AI collaborative, constrained iterative refinement process. This section presents its mathematical formalization, introduces a three-layer decoupled architecture, and models single refinement iterations via a finite state machine (FSM).

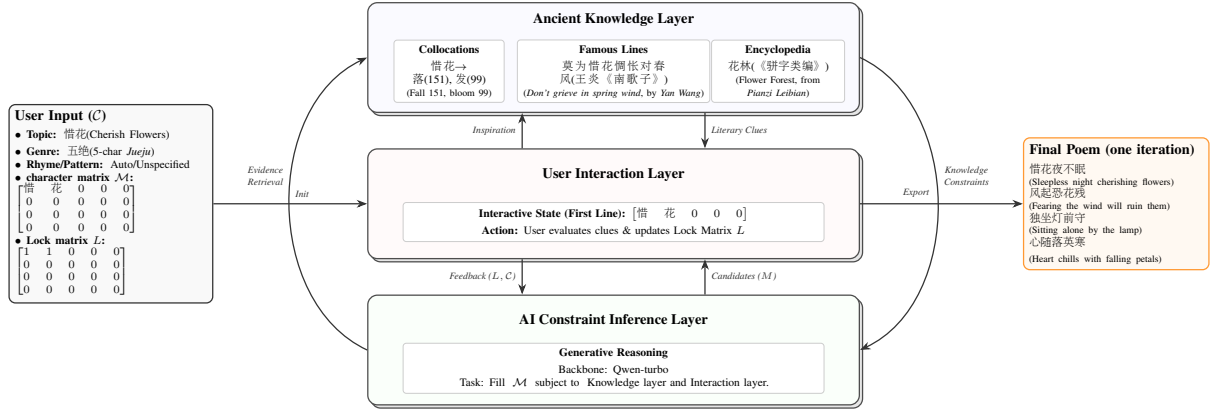
\begin{figure*}[t]
\centering
\begin{CJK*}{UTF8}{gbsn} 
\resizebox{\textwidth}{!}{
\begin{tikzpicture}[
    node distance=1.2cm,
    io node/.style={rectangle, draw=black!80, thick, fill=gray!5, text width=4.2cm, align=left, minimum height=2.4cm, font=\small, rounded corners=4pt},
    layer box/.style={rectangle, draw=black!70, thick, fill=white, text width=10.5cm, align=center, minimum height=2.4cm, rounded corners=6pt, font=\small\bfseries, copy shadow={shadow xshift=2pt, shadow yshift=-2pt, fill=black!10}},
    sub node/.style={rectangle, draw=gray!40, fill=white, font=\scriptsize, align=center, rounded corners=2pt, inner sep=3pt},
    arrow/.style={-{Stealth[scale=1.0]}, thick, draw=black!80},
    label font/.style={font=\tiny\itshape, color=black!80, align=center}
]

    \node[io node] (input) at (-4, -3.2) {
        \textbf{User Input ($\mathcal{C}$)} \\ \vspace{0.1cm}
        \scriptsize
        $\bullet$ \textbf{Topic:} 惜花 (Cherish Flowers) \\\vspace{0.1cm}
        $\bullet$ \textbf{Genre:} 五绝 (5-char \textit{Jueju}) \\
        $\bullet$ \textbf{Rhyme/Pattern:} Auto/Unspecified\\
        $\bullet$ \textbf{character matrix $\mathcal{M}$:} \\ $\begin{bmatrix} \text{惜} & \text{花} & 0 & 0 & 0 \\ 0 & 0 & 0 & 0 & 0 \\ 0 & 0 & 0 & 0 & 0 \\ 0 & 0 & 0 & 0 & 0 \end{bmatrix}$ \\
        $\bullet$ \textbf{Lock matrix $L$:}
        \\ $\begin{bmatrix} 1 & 1 & 0 & 0 & 0 \\ 0 & 0 & 0 & 0 & 0 \\0 & 0 & 0 & 0 & 0 \\0 & 0 & 0 & 0 & 0 \end{bmatrix}$
    };
    \node[layer box, fill=blue!2] (L1) at (7, 0) {};
    \node[anchor=north, font=\small\bfseries, yshift=-0.1cm] at (L1.north) {Ancient Knowledge Layer};
    
    \node[sub node, text width=2.2cm] (S1) at ($(L1.center)+(-3.6,-0.3)$) {
        \textbf{Collocations} \\ \vspace{0.1cm}
        惜花 $\rightarrow$ 落(151), 发(99) \\ \tiny (Fall 151, bloom 99)
    };
    \node[sub node, text width=4.0cm] (S2) at ($(L1.center)+(0,-0.3)$) {
        \textbf{Famous Lines} \\ \vspace{0.1cm}
        莫为惜花惆怅对春风(王炎《南歌子》) \\ \tiny (\textit{Don't grieve in spring wind}, by \textit{Yan Wang})
    };
    \node[sub node, text width=2.7cm] (S3) at ($(L1.center)+(3.6,-0.3)$) {
        \textbf{Encyclopedia} \\ \vspace{0.1cm}
        花林 (《骈字类编》) \\ \tiny (Flower Forest, from \textit{Pianzi Leibian})
    };
    \node[layer box, fill=red!2] (L2) at (7, -3.2) {};
    \node[anchor=north, font=\small\bfseries, yshift=-0.3cm] at (L2.north) {User Interaction Layer};
    
    \node[sub node, text width=9cm] (S4) at ($(L2.center)+(0,-0.4)$) {
        \textbf{Interactive State (First Line):} $\begin{bmatrix} \text{惜} & \text{花} & 0 & 0 & 0 \end{bmatrix}$ \\ \vspace{0.1cm}
        \textbf{Action:} User evaluates clues \& updates Lock Matrix $L$ \\
    };
    \node[layer box, fill=green!2] (L3) at (7, -6.4) {};
    \node[anchor=north, font=\small\bfseries, yshift=-0.3cm] at (L3.north) {AI Constraint Inference Layer};
    
    \node[sub node, text width=9cm] (S5) at ($(L3.center)+(0,-0.4)$) {
        \textbf{Generative Reasoning} \\ \vspace{0.1cm}
        Backbone: Qwen-turbo \\ 
        \text{Task: Fill } $\mathcal{M}$ \text{ subject to } Knowledge layer and Interaction layer.
    };
    \node[io node, fill=orange!5, draw=orange!80, text width=3.8cm] (output) at (18, -3.2) {
        \textbf{Final Poem (one iteration)} \\ \vspace{0.15cm} 
        \scriptsize
        \textbf{惜花夜不眠} \\ \tiny (Sleepless night cherishing flowers) \\
        \scriptsize
        \textbf{风起恐花残} \\ \tiny (Fearing the wind will ruin them) \\
        \scriptsize
        \textbf{独坐灯前守} \\ \tiny (Sitting alone by the lamp) \\
        \scriptsize
        \textbf{心随落英寒} \\ \tiny (Heart chills with falling petals)
    };
    \draw[arrow, transform canvas={xshift=-2cm}] (L2.north) -- (L1.south) node[midway, left, label font] {Inspiration};
    \draw[arrow, transform canvas={xshift=2cm}] (L1.south) -- (L2.north) node[midway, right, label font] {Literary Clues};

    \draw[arrow, transform canvas={xshift=-2cm}] (L2.south) -- (L3.north) node[midway, left, label font] {Feedback ($L, \mathcal{C}$)};
    \draw[arrow, transform canvas={xshift=2cm}] (L3.north) -- (L2.south) node[midway, right, label font] {Candidates ($M$)};

    \draw[arrow] (L1.east) to[out=-25, in=25] node[right, label font] {Knowledge\\Constraints\\ \ \\ \ \\ \ \\ \ \\ \ } (L3.east);
    \draw[arrow] (L3.west) to[out=155, in=205] node[left, label font] {Evidence\\Retrieval\\ \ \\ \  \\ \  \\ \ \\ \ } (L1.west);

    \draw[arrow] (input.east) -- (L2.west) node[midway, above, label font] {\ \ \ \ \ \ \ \ \ \ Init};
    \draw[arrow] (L2.east) -- (output.west) node[midway, above, label font] {Export\ \ \ \ \ \ \ \ \ \ \ \ \ \ \ \ \ \ \ \ \ };

\end{tikzpicture}
}
\end{CJK*} 
\caption{System architecture and the triadic collaborative iterative \textit{Tuiqiao} process. The core engine coordinates the Ancient Heritage, User Subjectivity, and AI Assistant layers through continuous feedback and evidence retrieval.}
\label{fig:system_arch}
\end{figure*}

\subsection{Problem Formulation}

We model a poem as a partially observable character matrix $M$, with dimensions determined by its poetic form (e.g., $M_{8\times5}$ for a 5-char \textit{lvshi}). Each entry $M_{i,j}$ contains either a Chinese character or a placeholder $0$ awaiting completion or refinement.

\textbf{Input Constraint Set $\mathcal{C}$.} 
The user defines the constraints as a set $\mathcal{C} = \{\mathcal{K}, d, \mathcal{Y}, \mathcal{P}, T\}$, comprising topic keywords $\mathcal{K}$, artistic description $d$, a target \textit{Pingshui Yun} rhyme category $\mathcal{Y}$, an optional first-line tonal pattern $\mathcal{P}$ (Appendix~\ref{sec:appendix}), and a verse-derived tonal matrix $T \in \{P,Z,A\}^{n\times m}$, where $P, Z, A$ denote level, oblique, and flexible tones, respectively.

\textbf{Interaction State \& Constraint Handling.}
The interaction state uses a current poem matrix $M_t$ and a lock matrix
$L \in \{0,1\}^{n\times m}$, where $L_{i,j}=1$ freezes a character and
$L_{i,j}=0$ keeps it editable. One \textit{Tuiqiao} step ranks returned
candidates using an implicit scoring function $\mathcal{S}$:
\begin{equation*}
\begin{aligned}
&M^*=\arg\max_M \mathcal{S}(M\mid M_t,L,\mathcal{C},E),\\
&\text{s.t. }M^*_{i,j}=(M_t)_{i,j},
   \quad \forall(i,j)\text{ where }L_{i,j}=1,\\
&\quad\ \text{Format}(M^*)=1,\\
&\quad\ \text{Rhyme}(M^*,\mathcal{Y})=1.
\end{aligned}
\end{equation*}
Here, $E$ denotes retrieved literary evidence. The formula characterizes
candidates returned after deterministic post-generation processing rather
than raw model outputs. Locked characters are restored and verified by
coordinates; candidates with invalid line or character counts are filtered
out; and rhyme violations are rejected, with validation feedback passed to
the next generation round for up to three rounds.

Tonal-pattern matching is handled separately and is not a hard acceptance
condition. The target pattern $T$ is included in the prompt and used to
prioritize better-matching candidates. Candidates with at most two tonal mismatches enter the preferred pool; if this pool is insufficient, basic candidates that pass the format, rhyme, and lock checks may still be returned with explicit mismatch warnings.
The user retains final authority to accept, reject, lock, or manually revise
every candidate.

\subsection{Triadic Collaborative Architecture}
We implement a three-layer decoupled architecture to support user-driven, ancient-guided, and AI-assisted collaboration (Figure~\ref{fig:system_arch}).

\textbf{Interaction Layer.} Captures fine-grained user operations (text input, character locking, suggestion selection) and translates them into system events. It dynamically updates the lock matrix $L$ and constraint set $\mathcal{C}$, maintaining full user control at each step.

\textbf{Knowledge Layer.} Grounded in classical scholarship, this layer provides literary justifications by integrating three types of knowledge (detailed in \S4): (1) high-frequency collocations extracted from 320,000 lines of Tang and Song poetry; (2) famous line references ranked via PPL using Qwen3-32B on 1.8M historical poems; and (3) 11 classical encyclopedias (e.g., \textit{Hailu Suishi}, \textit{Baikong Liutie}) supplying imagery, antithesis, and rhyme words.

\textbf{Constraint Inference Layer.} Powered by Qwen-turbo, this core engine aggregates hard constraints ($L, \mathcal{C}$) from the interaction layer and soft constraints (literary clues) from the knowledge layer. Through tailored prompt engineering and decoding strategies, it generates compliant poem candidates while positioning AI strictly as a background assistant.

\subsection{Iterative Refinement Cycle}
The continuous, user-led \textit{Tuiqiao} process is modeled as a finite state machine (FSM) loop: \\
(1) \textbf{Initialization}: The user inputs constraints $\mathcal{C}$ (and optionally a draft), and the system generates an initial poem matrix $M$. \\
(2) \textbf{AI Generation}: When the user triggers \textit{Tuiqiao}, the system populates placeholders in $M$ to produce candidate solutions. \\
(3) \textbf{Review \& Locking}: The user evaluates candidates and either locks satisfactory tokens (updating the lock matrix $L$) or manually edits the text. \\
(4) \textbf{Iterative Loop}: Steps (2) and (3) repeat dynamically with the updated $L$ until the user is satisfied. 

Throughout this cycle, the AI engine is strictly confined to generating interpretable options, while the knowledge layer supplies historical precedents as evidence. Ultimate creative agency and editing rights remain entirely with the human user, fully operationalizing our triadic collaborative philosophy.

%
%
\section{Methods}

To ensure interpretability, \textit{Jiuge-Tuiqiao} systematically integrates classical poetry knowledge with neural generation. For any refinement position, the system retrieves evidence from three knowledge sources: frequent co-occurrences, famous couplet matching, and ancient encyclopedia clues. Below, we detail the computation of each source and our joint fusion strategy.

\subsection{Frequent Co-occurrence Retrieval}
To discover habitual word pairings, we apply a $t$-test to assess collocation significance, filtering out spurious high-frequency pairings.

\textbf{Preprocessing \& Counting.} We utilize 320,000 lines from Tang-Song \textit{jueju} and \textit{lvshi} poems from Sou-Yun\footnote{\url{https://www.sou-yun.cn/}.}. Lines are segmented using a dictionary-plus-metrical-foot method (Appendix~\ref{sec:appendix_b}). After converting to simplified Chinese, filtering function words, and discarding short lines, we compute the global frequency $f(w)$ for each word $w$, and the co-occurrence frequency $f(w_i, w_j)$ for ordered pairs within the same line.

\textbf{Significance Testing.} Under the null hypothesis that words $w_i$ and $w_j$ appear independently within a line, their individual and joint probabilities are defined as $P(w) = f(w)/N$ and $P(w_i, w_j) = f(w_i, w_j)/N$, where $N$ is the total line count. The $t$-statistic is calculated as:
\begin{equation*}
T(w_i,w_j)=\frac{P(w_i,w_j)-P(w_i)P(w_j)}{\sqrt{P(w_i,w_j)/N}}.
\end{equation*}
We filter for pairs with $f(w_i,w_j)\ge3$ and $T>1.96$ (95\% confidence), sorting by descending $T$ to yield 28,739 high-frequency collocations.

\textbf{Suggestion Generation.} For a target position $(i,j)$ with context words $\mathcal{W}_{ctx}$, candidate characters $c$ are ranked by their highest $t$-value with any anchor word $w\in \mathcal{W}_{ctx}$, outputting the top candidates alongside their traceable collocation evidence.

\subsection{Famous Couplet Matching via Perplexity}
We employ Perplexity (PPL) as a proxy for canonical familiarity, assuming that lines with lower PPL are better predicted by a pretrained model and thus more likely to reflect canonical historical expressions. Low PPL is only a heuristic proxy: it may reflect canonical familiarity, pretraining memorization, or formulaic language. 

\textbf{Corpus Construction.} From 1.8M+ historical poems (Pre-Qin to Qing dynasty) sourced from Sou-Yun, we split texts by standard punctuation, filter out fragments shorter than 7 characters, and deduplicate to construct the candidate famous-line collection $\mathcal{D}_{raw}$.

\textbf{PPL Evaluation.} Using Qwen3-32B, we compute the PPL for each line $s=(x_1,...,x_{|s|})$ to obtain a model-based ranking score for retrieval:
\begin{equation*}
\text{PPL}(s)=\exp\left(-\frac{1}{|s|}\sum_{t=1}^{|s|}\log P(x_t\mid x_{<t};\theta)\right).
\end{equation*}
After ranking the lines by ascending PPL, we manually reviewed the top-ranked candidates and retained 20,000 verified lines to form the final famous-line index $\mathcal{D}_{FL}$.

\textbf{Binning \& Suggestion Generation.} To eliminate length bias (as longer lines inherently alter average PPL distributions), we stratify $\mathcal{D}_{FL}$ into distinct length bins using thresholds at 12 and 16 characters. For a target position $(i,j)$ and candidate character $c$, the system retrieves lines containing $c$ that overlap with the context $\mathcal{W}_{ctx}$. Candidates are then ranked by ascending PPL within their respective length bins and presented as curated, attributed historical examples.

\subsection{Ancient Encyclopedia Clue Retrieval}
We extract structured textual data from 11 ancient Chinese encyclopedias—including topic-organized works (e.g., \textit{Hailu Suishi}, \textit{Baikong Liutie}) and rhyme-organized works (e.g., \textit{Yunfu Qunyu})—to construct three specialized databases (detailed in Appendix~\ref{sec:appendix_c}):\\ 
(1) \textbf{Imagery Words}, storing topic-specific poetic images and descriptions;\\
(2) \textbf{Antithesis Words}, containing parallel or antithetical word pairs; \\
(3) \textbf{Rhyme Words}, categorizing terms by their prescriptive rhyme categories.

\textbf{Suggestion Generation.} For a target position $(i,j)$ given the current word $w$ and user keywords $\mathcal{K}$, the system dynamically queries these databases. It retrieves relevant imagery clues based on $\mathcal{K}$, antithesis clues based on $w$ (if parallel constraints apply), or rhyme clues matching the target rhyme $\mathcal{Y}$. These retrieved entries are returned directly to the user as explicit literary justifications.

\subsection{Suggestion Organization and Presentation}

The interface displays the suggestion lists from three knowledge sources in parallel as categorized cards. Users can browse this traceable evidence to either adopt a candidate or perform manual edits, preserving ultimate human agency over the final text.

%
\section{User Interface}
The \textit{Jiuge-Tuiqiao} WeChat mini-program (Taro 4, React, FastAPI) features a human-centric interface that balances direct user control with multi-channel poetic knowledge.

\begin{figure}[t]
    \centering
    \includegraphics[width=\linewidth]{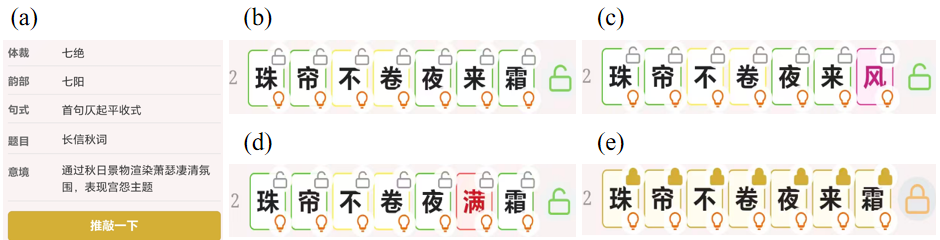}
    \caption{Overview of system states: (a) metadata constraint initialization panel; (b--e) real-time feedback in the Character Grid Editor showing (b) valid compliance (green) and acceptable variants (yellow), (c) rhyme mismatch (purple), (d) tonal violation (red), and (e) user-locked tokens (gold fill).}
    \label{fig:ui1}
\end{figure}

\textbf{Character Grid Editor.} As the central workspace, this grid dynamically renders characters alongside multi-dimensional states (Figure~\ref{fig:ui1}). With a 500\,ms debounce, it provides real-time feedback: \textbf{green} borders denote valid compliance and \textbf{yellow} marks acceptable variants (Figure~\ref{fig:ui1}b); \textbf{purple} signals rhyme mismatches (Figure~\ref{fig:ui1}c); and \textbf{red} indicates tonal violations (Figure~\ref{fig:ui1}d). A \textbf{gold fill} highlights user-locked tokens (Figure~\ref{fig:ui1}e). Explicit justifications accompany violations to streamline error correction.

\textbf{Hierarchical Locking Mechanism.} To preserve human creative agency, users can restrict AI generation at two granularities: (1) \textit{Character-level lock} freezes individual cells ($L_{i,j} \leftarrow 1$, gold fill; Figure~\ref{fig:ui1}e) to preserve user tokens; (2) \textit{Line-level lock} fixes an entire row, forcing the system to generate alternatives strictly for the remaining lines.

\begin{figure}[t]
    \centering
    \includegraphics[width=1.0\linewidth]{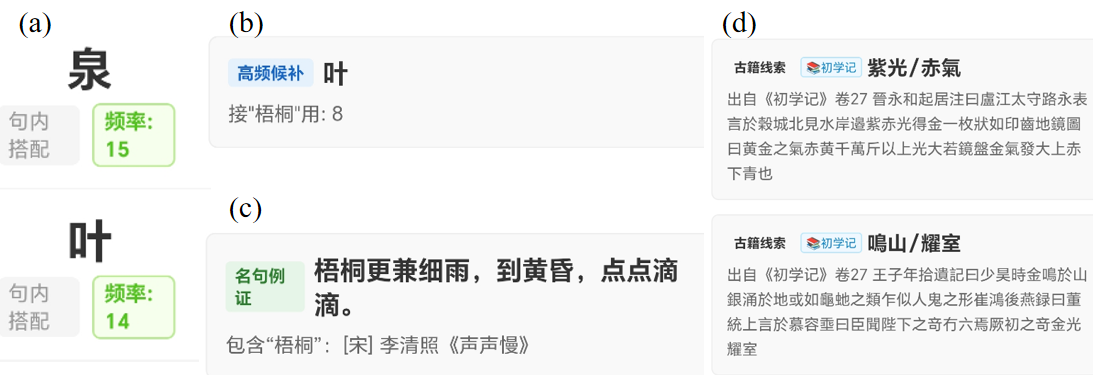}
    \caption{Multi-channel evidence in the Interpretable \textit{Tuiqiao} Panel: (a--b) statistical high-frequency co-occurrence candidates, (c) ranked famous-line samples with metadata, and (d) textual clues retrieved from ancient encyclopedias.}
    \label{fig:ui2}
\end{figure}

\textbf{Interpretable \textit{Tuiqiao} Panel.} Clicking an unlocked cell activates a side panel that first aggregates evidence from three provenance-backed channels (Figure~\ref{fig:ui2}): (1) the \textit{High-frequency co-occurrence} channel displays top collocates from a 28,739-entry table (Section~4.1, Figure~\ref{fig:ui2}(a--b)); (2) the \textit{Famous-line samples} channel displays up to five lines ranked by ascending PPL with full source metadata (Section~4.2, Figure~\ref{fig:ui2}(c)); and (3) the \textit{Ancient encyclopedia clues} channel provides imagery, antithesis, or rhyme clues mined from 11 historical encyclopedias (Section~4.3, Figure~\ref{fig:ui2}(d)). When none of these sources yields a match, the system invokes an LLM to generate a fallback rationale, which is served asynchronously through an 8,000-capacity LFU cache to reduce latency. This fallback is explicitly presented as model-generated and is not treated as provenance-backed literary evidence. Because the in-system famous-line index is limited to 20,000 lines retained after ranking by ascending PPL and manual verification, the active \textit{Sou-Yun} integration hyperlinks each grid character to the corresponding search results, allowing users to inspect the full set of relevant historical poetic lines indexed by \textit{Sou-Yun}. This extends evidence inspection beyond the local index while leaving the final decision to the poet.

\textbf{Revision History \& Workflow.} Each line maintains an independent undo/redo stack for single-click version restoration. This supports an iterative co-creation loop: users initialize metadata (Figure~\ref{fig:ui1}a), generate candidates, lock satisfactory segments, and progressively refine tokens via the \textit{Tuiqiao} panel under real-time validation until completion.

%
\section{Evaluation}
We evaluate Jiuge-Tuiqiao across three dimensions:
(1) a knowledge-source ablation study on cloze-filling;
(2) automatic prosodic metrics on generated poems; and
(3) a human evaluation assessing creative experiences.

\subsection{Ablation Study: Famous-Line and Co-occurrence Contributions}
\textbf{Task and Setup.} To assess the individual and combined contributions of the famous-line and co-occurrence retrieval sources, we design a controlled cloze-filling task using 50 poem lines stratified by poem type. For each instance, we mask a contiguous two-character content word and prompt the backbone LLM (Qwen-turbo) to predict the blank under four conditions: (1) \textbf{BASE}: context only; (2) \textbf{+FL}: adds up to five PPL-ranked famous lines containing the anchor term; (3) \textbf{+CO}: adds top-20 co-occurrence phrases; and (4) \textbf{BOTH}: provides both sources simultaneously. For each instance, we excluded the original source line from the famous-line index before retrieval. Then, we evaluate performance using \textbf{Hit@1} (whether the original ground-truth word is the top-ranked candidate) and \textbf{Hit@5} (whether the ground truth appears anywhere within the top five suggestions).

\textbf{Results.} As shown in Table~\ref{tab:ablation}, auxiliary knowledge largely boosts performance over the BASE condition. While \textbf{+FL} yields the highest Hit@1 (60\%) due to historical exact-matches, introducing co-occurrences (\textbf{+CO} and \textbf{BOTH}) expands the lexical search space. Notably, \textbf{BOTH} maintains the peak Hit@5 (60\%) while slightly softening Hit@1 to 56\%. This trade-off is ideal for an interactive assistant: famous lines preserve classical diction, while co-occurrence phrases inject stylistic diversity rather than merely forcing the reproduction of the ground truth.

\begin{table}[h]
  \centering
  \small
  \begin{tabular}{lcc}
    \hline
    \textbf{Condition} & \textbf{Hit@1} & \textbf{Hit@5} \\
    \hline
    BASE & 12\% & 12\% \\
    +FL  & 60\% & 60\% \\
    +CO  & 36\% & 42\% \\
    BOTH & 56\% & 60\% \\
    \hline
  \end{tabular}
  \caption{Ablation results on cloze-filling. Combining both sources (\textbf{BOTH}) matches the highest Hit@5 (60\%) while yielding a slightly lower Hit@1 (56\%) than \textbf{+FL} alone (60\%).}
  \label{tab:ablation}
\end{table}

\subsection{Automatic Prosody Evaluation}
\textbf{Setup \& Metrics.} We evaluate system-generated poems using 80 requests stratified across four forms (5/7-character \textit{jueju} and \textit{lvshi}) and four historical title-frequency tiers: \textit{high} (top 10\%), \textit{mid} (10--50\%), \textit{low} (50--90\%), and \textit{rare} (bottom 10\%). We report rule-based correctness via \textbf{Format} (structural length), \textbf{L-Pattern} (line-level tonal accuracy), \textbf{P-Pattern} (global poem-level pattern adherence including \textit{nian-dui} constraints), and \textbf{Rhyme} (\textit{Pingshui Yun} consistency). Output diversity is quantified using \textbf{Distinct-$n$} and \textbf{Self-BLEU} ($n \in \{1,2,4\}$).

\textbf{Results.} As shown in Table~\ref{tab:prosody}, the system achieves perfect Format and Rhyme accuracy across all forms, alongside high line-level alignment (L-Pattern $\ge 91.25\%$). However, global constraints impose a strict bottleneck: poem-level adherence (P-Pattern) drops significantly for \textit{lvshi} ($15\%$--$25\%$) compared to \textit{jueju} ($50\%$--$65\%$), reflecting the complexity of compounding eight-line \textit{nian-dui} requirements. Crucially, Table~\ref{tab:diversity} confirms that these structural constraints do not induce mode collapse; lexical diversity remains high, with Distinct-4 reaching 1.0000 and Self-BLEU-4 at a low 0.1855.

\begin{table}[h]
  \centering
  \small
  \setlength{\tabcolsep}{4pt}
  \begin{tabular}{lcccc}
    \hline
    \textbf{Form} & \textbf{Format} & \textbf{L-Pattern} & \textbf{P-Pattern} & \textbf{Rhyme} \\
    \hline
    5-char \textit{jueju} & 100\% & 96.25\% & 65\% & 100\% \\
    7-char \textit{jueju} & 100\% & 91.25\% & 50\% & 100\% \\
    5-char \textit{lvshi} & 100\% & 93.12\% & 25\% & 100\% \\
    7-char \textit{lvshi} & 100\% & 93.12\% & 15\% & 100\% \\
    \hline
  \end{tabular}
  \caption{Automatic prosody accuracy by verse form.}
  \label{tab:prosody}
\end{table}

\begin{table}[h]
  \centering
  \small
  \begin{tabular}{lccc}
    \hline
                   & \textbf{1-gram} & \textbf{2-gram} & \textbf{4-gram} \\
    \hline
    Distinct   & 0.9804 & 0.9995 & 1.0000 \\
    Self-BLEU  & 0.8983 & 0.5717 & 0.1855 \\
    \hline
  \end{tabular}
  \caption{Output diversity metrics across generations.}
  \label{tab:diversity}
\end{table}

\subsection{User Evaluation}
\label{sec:user}

\paragraph{Setup.}
We conducted a preliminary user evaluation with 10 participants (poetry enthusiasts and researchers in classical Chinese literature). After composing at least one poem, each participant completed a 9-point Likert scale questionnaire (1 = strongly disagree, 9 = strongly agree) covering four dimensions: (1) \textbf{Agency} (active creative control), (2) \textbf{Trustworthiness} (suggestion credibility), (3) \textbf{Prosody guidance} (color feedback utility), and (4) \textbf{Overall satisfaction}. Open-ended qualitative feedback was also collected.

\begin{figure}[t]
  \centering
  \includegraphics[width=.95\columnwidth]{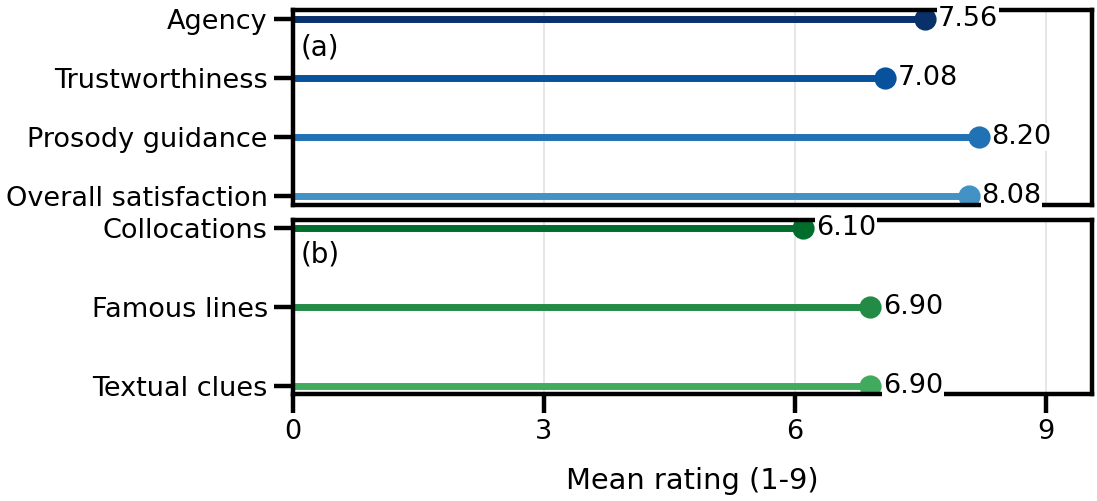}
  \caption{Human evaluation ratings (9-point Likert scale) showing (a) four evaluation dimensions and (b) perceived utility of three knowledge sources.}
  \label{fig:user_results}
\end{figure}

\paragraph{Results.}
As shown in Figure~\ref{fig:user_results}, participants rated the system positively overall. Prosody guidance scored highest (8.20/9), demonstrating that real-time color feedback effectively aided metrical revisions. Overall satisfaction (8.08/9) and agency (7.56/9) were high, confirming active user control over AI outputs. Trustworthiness and evidence utility scored 7.08/9. Among knowledge sources, users preferred famous lines and ancient textual clues over frequent collocations, prioritizing contextual literary evidence over statistical signals. Open-ended feedback highlighted that evidence panels (source lines, author attributions, citations) enhanced credibility and facilitated in-context learning.

\section{Conclusion and Future Work}

We presented \textit{Jiuge-Tuiqiao}, a human-AI collaborative poetry composition system that re-centers human agency through classical aesthetics. The framework implements a triadic design—integrating hierarchical locking, real-time prosody validation, and multi-channel traceable recommendations. Its central contribution is to operationalize interpretability as provenance at the interface level: for retrieved suggestions, users can inspect the supporting statistical or literary sources while retaining final authority to accept, reject, lock, or manually revise the text.

Future work will focus on four directions: 
(1) \textbf{Personalized modeling} to adapt recommendations to individual stylistic preferences across sessions; 
(2) \textbf{Broader poetic forms} extending constraints to irregular meters like \textit{ci} (song-lyrics) and \textit{sanqu} (aria); 
(3) \textbf{Comprehensive user studies} engaging larger, diverse participant pools to rigorously validate the system's usability and long-term creative impact; and
(4) \textbf{Preference alignment} utilizing interaction logs as training signals to fine-tune LLMs toward both high poetic quality and user agency.

\section*{Ethics and Impact}

\textbf{Creative authorship and attribution.} \textit{Jiuge-Tuiqiao} is designed so that the human poet retains full creative initiative: the system provides options and evidence, never unilaterally inserting text without user acceptance. Nevertheless, AI-assisted composition raises genuine questions about authorship. We recommend that users who publish AI-assisted poems disclose the assistance; the system's export functionality includes a disclosure template for this purpose. We do not claim that AI-generated candidates carry the same creative authorship as wholly human-written verse.

\textbf{Cultural preservation and accessibility.} Classical Chinese poetry encodes millennia of cultural knowledge — prosodic conventions, canonical imagery, allusive traditions. By embedding the \textit{Pingshui Yun} system, ancient encyclopedias, and a vast famous-line corpus into the interaction loop, \textit{Jiuge-Tuiqiao} lowers the barrier to engaging with and practising this tradition, potentially reaching learners of Chinese as a second language and heritage communities outside China.

\textbf{Potential biases.} The knowledge base is constructed from historical corpora that skew heavily toward the Tang and Song dynasties and toward elite male authorship. As a result, recommended collocations and famous lines may systematically under-represent the poetic traditions of women poets, border literati, and minority-language contributors to the Chinese literary canon. Future versions should apply diversity-aware retrieval to broaden the representational range of suggestions.

\textbf{Risk of creative dependency.} A persistently available refinement tool may, over time, discourage users from developing independent prosodic intuition. The system mitigates this risk by displaying explanatory feedback for each prosodic violation rather than silently correcting it, and by making all knowledge sources and their justifications explicitly visible — encouraging users to learn \textit{why} a suggestion is grounded in classical practice, not merely to accept it.

\section*{Acknowledgments}
We thank the anonymous reviewers for their constructive comments. We are also grateful to the volunteers who participated in the user evaluation for their ratings and valuable feedback on the system. This work was supported by the National Natural Science Foundation of China (Grant No. 62236011), the Key Laboratory of Ethnic Language Intelligent Analysis and Security Governance of the Ministry of Education at Minzu University of China, Beijing, China, and the Shenzhen--Tsinghua Special Program for Fundamental and Frontier Research in Artificial Intelligence, under the project ``Development of an AI-Driven Next-Generation Digital Humanities Research Platform'' (No. AI2026027).


\bibliography{custom}

\clearpage
\appendix
\section{Rhyme Categories and First-line Tonal Patterns}
\label{sec:appendix}
\begin{CJK*}{UTF8}{gbsn}

\textbf{Pingshui Yun Rhyme Categories (30 Level Tones).} The system supports 30 Level Tone categories from the \textit{Pingshui Yun} system. 

\textbf{Upper Level Tones (上平声):} 东Dong, 冬Dong, 江Jiang, 支Zhi, 微Wei, 鱼Yu, 虞Yu, 齐Qi, 佳Jia, 灰Hui, 真Zhen, 文Wen, 元Yuan, 寒Han, 删Shan. 

\textbf{Lower Level Tones (下平声):} 先Xian, 萧Xiao, 肴Yao, 豪Hao, 歌Ge, 麻Ma, 阳Yang, 庚Geng, 青Qing, 蒸Zheng, 尤You, 侵Qin, 覃Tan, 盐Yan, 咸Xian.

\vspace{1.5ex}
\textbf{Tonal Pattern Matrices (7-character \textit{Jueju}).} Tonal constraints are modeled as matrices $T \in \{P, Z, A\}^{n \times m}$, where $P$: Level tone (平声), $Z$: Oblique tone (仄声), and $A$: Flexible. The four globally valid patterns are consolidated below.

{\small
\begin{gather*}
\text{1. Level-start, Level-end (平起平收)} \\
T_{L-L} = \begin{bmatrix} A & P & A & Z & Z & P & P \\ A & Z & P & P & A & Z & P \\ A & Z & A & P & P & Z & Z \\ A & P & A & Z & Z & P & P \end{bmatrix} \\
\text{2. Level-start, Oblique-end (平起仄收)} \\
T_{L-O} = \begin{bmatrix} A & P & A & Z & A & P & Z \\ A & Z & P & P & A & Z & P \\ A & Z & A & P & P & Z & Z \\ A & P & A & Z & Z & P & P \end{bmatrix} \\
\text{3. Oblique-start, Level-end (仄起平收)} \\
T_{O-L} = \begin{bmatrix} A & Z & P & P & A & Z & P \\ A & P & A & Z & Z & P & P \\ A & P & A & Z & A & P & Z \\ A & Z & P & P & A & Z & P \end{bmatrix} \\
\text{4. Oblique-start, Oblique-end (仄起仄收)} \\
T_{O-O} = \begin{bmatrix} A & Z & A & P & P & Z & Z \\ A & P & A & Z & Z & P & P \\ A & P & A & Z & A & P & Z \\ A & Z & P & P & A & Z & P \end{bmatrix}
\end{gather*}
}
\end{CJK*}

\section{Poem Segmentation Algorithm}
\label{sec:appendix_b}

\textbf{Two-Stage Segmentation Strategy.} The system employs a two-stage strategy to achieve grammar-based segmentation (G-Seg): (1) \textit{Candidate Generation}: Based on a scoring dictionary and metrical rules, the system enumerates all possible paths via Depth-First Search (DFS) under maximum word length limits, globally selecting the top-$K$ scoring candidate paths. (2) \textit{LLM Reranking}: An LLM acts as a discriminator to select the optimal candidate that best adheres to grammatical norms from the top-$K$ list.

\textbf{Dictionary Scoring Formula.} The total score for a segmentation path is defined as $\text{Score} = \sum_{w_i \in \text{path}} \text{Score}_{dict}(w_i) + \text{Bonus}_{meter}(w_i)$. The base dictionary score $\text{Score}_{dict}(w)$ for a word $w$ of length $L$ integrates multiple empirical lexical features:
\begin{equation*}
\begin{split}
\text{Score}_{dict}(w)& = \\
& \log(1 + \text{freq}(w)) + 0.8 \cdot \text{MI}_{norm}(w) \\
&+ 0.3 \cdot \text{DictCount}(w) + \text{Allusion}(w) \\
&+ \text{LenPref}(L) - \text{OOV\_Penalty},
\end{split}
\end{equation*}
where $\text{MI}_{norm}$ represents normalized mutual information, and $\text{DictCount}$ tracks occurrence counts across historical dictionaries. $\text{Allusion}(w)$ adds $+1.0$ for established literary references. The length preference term is defined as $\text{LenPref}(L) = \{+0.5 \text{ for } L=2, +0.2 \text{ for } L=3, -0.2 \text{ for } L \ge 4\}$. Out-of-vocabulary (OOV) tokens receive a base score of $-1.0$ and an additional penalty of $-0.1$.

\begin{table}[t]
\centering
\small 
\begin{tabular}{lll}
\hline
\textbf{Line Type} & \textbf{Boundary Rules} & \textbf{Length Rules} \\ \hline
\multirow{2}{*}{5-char} & Pos 2: $+0.8$ & 2-char: $+0.1$ \\
 & Pos 3: $-0.3$ & 1-char: $-0.05$ \\ \hline
\multirow{3}{*}{7-char} & Pos 2, 4: $+0.5$ & 2-char: $+0.1$ \\
 & Pos 3, 5: $-0.3$ & 1-char: $-0.05$ \\
 & & $\ge 4$-char: $-0.1$ \\ \hline
\multirow{2}{*}{General} & \multirow{2}{*}{-} & 2-char: $+0.05$ \\& &1-char: $-0.05$ \\ \hline
\end{tabular}
\caption{\label{tab:metrical_rhythm}Metrical rhythm reward and penalty rules.}
\vspace*{-3ex}
\end{table}

\textbf{Metrical Constraints.} To simulate natural classical Chinese recitation rhythms ($2+3$ for 5-character and $2+2+3$ for 7-character lines), position- and length-specific bonuses ($\text{Bonus}_{meter}$) are applied directly following the operational rules defined in Table~\ref{tab:metrical_rhythm}.

\textbf{G-Seg Specification for LLM Reranking.} In Stage 2, the LLM evaluates the top-$K$ candidates via few-shot prompts under strict linguistic guidelines: (1) functional words (e.g., particles, adverbs, conjunctions) must stand alone as single characters whenever possible; (2) tightly coupled semantic entities (names like ``Huang Siniang'', places) must be merged into single tokens; (3) contextual flexibility takes precedence over blindly preferring longer words. If no available candidate is grammatically sound, the LLM triggers a \texttt{WARNING} fallback and outputs its self-corrected proper segmentation sequence.

\begin{CJK*}{UTF8}{gbsn}
\begin{table*}[!t]
\centering
\scriptsize
\renewcommand{\arraystretch}{1.5}
\begin{tabular}{p{2.2cm} p{4.8cm} p{8.5cm}}
\hline
\textbf{Book} & \textbf{Description} & \textbf{Content Example (Extracted Data)} \\ \hline
\textbf{《北堂书钞》}\newline \textit{Beitang Shuchao} & Compiled by Yu Shinan (Tang). An early extant encyclopedia focusing on politics and rituals, preserving numerous pre-Sui texts. & \textbf{Keyword:} 功业 (Achievements) \  \textbf{Volume:} 4 \newline  \textbf{Section:} 帝王 (Emperors) \newline \textbf{Content:} ``四本具即帝初立，举而措之事业，功业赫赫，功盛德厚，功侔太古...'' \\ \hline

\textbf{《白孔六帖》}\newline \textit{Baikong Liutie} & Compiled by Bai Juyi (Tang) and Kong Chuan (Song). Collects phrases and sentences as source materials for poetry and essays.& \textbf{Keyword:} 天 (Heaven) \newline \textbf{Volume:} 1 \newline \textbf{Content:} ``(白)高明柔克(髙明天也柔克寒暑不干), 阴骘下民(言天黙定下民之命)...'' \\ \hline

\textbf{《太平御览》}\newline \textit{Taiping Yulan} & A massive state-sponsored encyclopedia from the early Song Dynasty. It is highly comprehensive, systematically categorized by heaven, earth, human, objects, etc. & \textbf{Keyword:} 太初 (Vital Energy) \ \textbf{Volume:} 1 \newline \textbf{Section:} 天部一 (Heaven) \newline \textbf{Content:} ``《易乾鉴度》曰：太初者，气之始也。《帝王世纪》曰：元气始荫，谓之太初...''  \\ \hline

\textbf{《艺文类聚》}\newline \textit{Yiwen Leiju} & Compiled by Ouyang Xun (Tang). A state-sponsored encyclopedia intertwining historical facts and literature (poems and prose) under a clear Confucian orthodox viewpoint. & \textbf{Keyword:} 日 (Sun) \ \textbf{Volume:} 1 \newline \textbf{Ref Word:} 天部上 (Heaven) \newline \textbf{Content:} ``《易》曰：日月丽乎天。又曰：离为日。又曰：日中则昃，月盈则食。天地盈虚，与时消息。而况于人乎，况于鬼神乎...'' \\ \hline

\textbf{《初学记》}\newline \textit{Chuxueji} & Compiled by Xu Jian (Tang). An introductory encyclopedia strictly organized by chapters (heaven, earth, etc.), drawing from historical literature and preserving ancient fragments. & \textbf{Keyword:} 日 (Sun) \ \textbf{Volume:} 1 \ \textbf{Section:} 天部(Heaven) \newline \textbf{Content:} 叙事(Narrative): "说文云日者实也...";\  事对(Events): 丽天\&出地（易曰日月丽乎天百谷草木丽乎地文子曰日出于地...;\  诗文(Poetry): "梁简文帝咏朝日诗（团团出天外煜煜上层峰光随浪高下影逐树轻浓）..."\\ \hline

\textbf{《骈字类编》}\newline \textit{Pianzi Leibian} & Compiled by Shen Zongjing et al. under imperial order (Qing). It collects disyllabic compound words (\textit{Pianzi}) with sources for phrasing reference.  & \textbf{Keyword:} 天地 (Heaven and Earth) \ \textbf{Volume:} 1 \newline \textbf{Section:} 天 (Heaven) \newline \textbf{Content:} ``易干夫大人者与天地合其德，又坤天地变化草木蕃天地闭贤人隐...'' \\ \hline

\textbf{《海录碎事》}\newline \textit{Hailu Suishi} & Compiled by Ye Tinggui (Song). Supplements missing allusions from larger encyclopedias with concise entries of Tang poems and prose. & \textbf{Keyword:} 天末 (Horizon) \ \textbf{Volume:} 1 \newline \textbf{Section:} 天(Heaven) \newline \textbf{Content:} ``沧波眇川汜白日隐天末（李白诗）...'' \\ \hline

\textbf{《佩文韵府》}\newline \textit{Peiwen Yunfu} & An imperially compiled Qing rhyme dictionary arranged by 106 \textit{Pingshui Yun} categories. It details characters in each group and provides phonological analysis for writers. & \textbf{Keyword:} 东 (East) \ \textbf{Volume:} 01之一 \ \textbf{Tones:} 上平声(Shangping) \newline \textbf{Rhyme:} 一东(Dong)(东德红切眷方也...) \newline \textbf{Words of Rhyme:} 自东(诗我来自东又自西自东) 大东(诗遂荒大东)... \newline \textbf{Words of Pairs:} 渭北\&江东，日下\&天东，河内\&济东... \newline \textbf{Excerpt:} 力障百川东，光升必自东...\\ \hline

\textbf{《声律启蒙》}\newline \textit{Shenglü Qimeng} & A Qing rhyme primer by Che Wanyu focusing on antithesis training.  & \textbf{Antithesis Pairs:} ``云'' (Cloud) vs. ``雨'' (Rain); ``雪'' (Snow) vs. ``风'' (Wind) \\ \hline

\textbf{《笠翁对韵》}\newline \textit{Liweng Duiyun} & A Qing rhyme primer by Li Yu for antithesis and allusion training.  & \textbf{Antithesis Pairs:} ``天'' (Heaven) vs. ``地'' (Earth); ``雨'' (Rain) vs. ``风'' (Wind) \\ \hline

\textbf{《龙文鞭影》}\newline \textit{Longwen Bianying} & A Ming primer by Xiao Liangyou utilizing four-character sentences to catalog historical names, places, and allusions. & \textbf{Antithesis Pairs:} ``粗成四字'' vs. ``诲尔童蒙'' \\ \hline

\end{tabular}
\caption{Overview and data samples of the 11 utilized classical sourcebooks.}
\label{tab:leishu_overview}
\vspace*{-2ex}
\end{table*}
\section{Ancient Encyclopedias Database} 
\label{sec:appendix_c}

To provide well-grounded philological and linguistic suggestions during the \textit{Tuiqiao} process, our system constructs a structured knowledge base by extracting textual data from over ten representative classical Chinese encyclopedias (\textit{Leishu}) and rhyme dictionaries spanning the Tang to Qing dynasties (e.g., \textit{Chuxueji}, \textit{Taiping Yulan}, and \textit{Peiwen Yunfu}). Table~\ref{tab:leishu_overview} outlines the basic profiles and concrete database entry examples of these primary resources.

\vspace{1.5ex}
\textbf{Data Processing and Knowledge Base Construction.} Texts from the \textit{Siku Quanshu} editions and verified open-source assets were systematically parsed, cleaned, and rule-filtered to unify disparate historical formats. The consolidated pipeline yields 503,908 structured records written across three distinct SQLite index tables to serve as explicit etymological anchors:

\textbf{(1) Image Word Database (\texttt{tb\_danci}):} Aggregates 6 sourcebooks (e.g., \textit{Beitang Shuchao}, \textit{Taiping Yulan}) to map a central thematic keyword (\texttt{key\_word}) to extensive imagery tokens historically favored by poets, attaching original definitions and cited literature contexts (\texttt{content}) as conceptual justifications during theme refinement.

\textbf{(2) Antithesis Word Database (\texttt{tb\_duizhang}):} Combines 5 phonological books (e.g., \textit{Chuxueji}, \textit{Shenglü Qimeng}) by extracting all original antithetical couplets (\texttt{gen\_word\_f} and \texttt{gen\_word\_b}) into a parallel-pair database. The system dynamically recommends position-specific antithetical words calibrated against the corresponding characters in parallel lines, returning classical source texts as reference.

\textbf{(3) Rhyme Word Database (\texttt{tb\_yunyu}):} Specially processes texts structured around rhyme schemas (primarily \textit{Peiwen Yunfu}). It isolates the leading characters of \textit{Pingshui Yun} categories (\texttt{yun\_word}) and their corresponding phrases (\texttt{gen\_word}) to supply rhyme-compliant generation candidates.

Ultimately, these verified text streams are not only injected into the language model as internal structure constraints (Constraint Inference Layer) but are also rendered explicitly to users via interactive cards, securing high interpretability and strict knowledge tracing for \textit{Jiuge-Tuiqiao} framework.

\end{CJK*}

\end{document}